\documentclass[letterpaper, 10 pt, conference]{ieeeconf}  

\IEEEoverridecommandlockouts                              

\usepackage{graphicx} 
\usepackage{amsmath} 
\usepackage{amssymb}  
\usepackage{color, soul}
\usepackage[lofdepth,lotdepth]{subfig}

\usepackage{tikz}

\newcommand{\tikzcirclegreen}[2][green,fill=white]{\tikz[baseline=-0.7ex]\draw[ultra thick,#1,radius=#2] (0,0) circle (0.08cm);}%
\newcommand{\tikzcirclered}[2][red,fill=white]{\tikz[baseline=-0.7ex]\draw[ultra thick,#1,radius=#2] (0,0) circle (0.08cm);}%
\newcommand{\tikzrectangleblue}[2][blue,fill=white]{\tikz[baseline=-0.1ex]\draw[thick,#1,radius=#2] (0,0) rectangle (0.3cm, 0.2cm);}%
\newcommand{\tikzrectanglegreen}[2][green,fill=white]{\tikz[baseline=-0.1ex]\draw[thick,#1,radius=#2] (0,0) rectangle (0.3cm, 0.2cm);}%
\newcommand{\bz}{\mathbf{z}}
\newcommand{\bp}{\mathbf{p}}

\newcommand{\bbR}{\mathbb{R}}
\newcommand*{\mathcolor}{}
\def\mathcolor#1#{\mathcoloraux{#1}}
\newcommand*{\mathcoloraux}[3]{%
	\protect\leavevmode
	\begingroup
	\color#1{#2}#3%
	\endgroup
}
\definecolor{myred}{RGB}{255,0,0}
\definecolor{mygreen}{RGB}{0,255,0}

\title{\LARGE \bf
	Learning to Predict Repeatability of Interest Points
}

\author{Anh-Dzung Doan$^{1}$, Daniyar Turmukhambetov$^{2}$, Yasir Latif$^{1}$, Tat-Jun Chin$^{1}$, and Soohyun Bae$^{2}$
\thanks{*This work was done during an internship at Niantic.}
\thanks{$^{1}$Anh-Dzung Doan, Yasir Latif, and Tat-Jun Chin are with School of Computer Science, The University of Adelaide.}
\thanks{$^{2}$Daniyar Turmukhambetov and Soohyun Bae are with Niantic.}}

\begin{document}

\maketitle
\thispagestyle{empty}
\pagestyle{empty}

\begin{abstract}
    Many robotics applications require interest points that are highly repeatable under varying viewpoints and lighting conditions. However, this requirement is very challenging as the environment changes continuously and indefinitely, leading to appearance changes of interest points with respect to time. This paper proposes to predict the repeatability of an interest point as a function of time, which can tell us the lifespan of the interest point considering daily or seasonal variation. The repeatability predictor (RP) is formulated as a regressor trained on repeated interest points from multiple viewpoints over a long period of time. Through comprehensive experiments, we demonstrate that our RP can estimate when a new interest point is repeated, and also highlight an insightful analysis about this problem. For further comparison, we apply our RP to the map summarization under visual localization framework, which builds a compact representation of the full context map given the query time. The experimental result shows a careful selection of potentially repeatable interest points predicted by our RP can significantly mitigate the degeneration of localization accuracy from map summarization\footnote{We will make the source code publicly available}.
\end{abstract}

\section{INTRODUCTION}

Local interest points of images play a vital role in a wide range of robotic vision applications, e.g., visual SLAM~\cite{monoslam,orbslam2, baslam,kimera}, place recognition~\cite{fabmap,bagofbinarywords,doan2019scalable}, visual localization~\cite{experiencebasednavigation, paton2016bridging, doan2020visual}, change detection~\cite{palazzolo2018fast,ulusoy2014image, taneja2011image}, etc. Its sparsity provides many advantages including an efficient memory storage yet effective correspondence estimation, so it is favored in large-scale applications~\cite{lynen2020large,tran2018device,lynen2015get}. To explore the properties of interest points, Schmid et al.~\cite{schmid98repeatability} introduced \textit{repeatability} to characterize if a 3D scene point detected in the first image can also be detected in the second one. However, in practice, repeatability needs to be associated with \textit{matchability}, which identifies if two interest points correspond to a single 3D point~\cite{hartmann2014predicting}. Hence, the number of repeatable interest points are the upper bound of number of correspondences found via feature matching~\cite{zhou2016evaluating}.

In literature, a considerable effort has been made to find a local interest point detector robust against environmental changes \cite{sift, r2d2, d2net}, whose basic strategy is to detect interest points with high repeatability scores. However, these approaches make a strong assumption, i.e., interest points must be \textit{repeatable} regardless of environmental changes w.r.t time span. In practice, according to \cite{stylianou2015characterizing}, this assumption does not hold because environmental changes (including physical, weather, and illumination changes) affect to the physical appearance of 3D points. Consequently, \cite{stylianou2015characterizing} shows a decay in terms of the feature matching performance, which is crucially caused by the degradation of repeatability.

\begin{figure}
	\centering
	\includegraphics[width=0.45\textwidth]{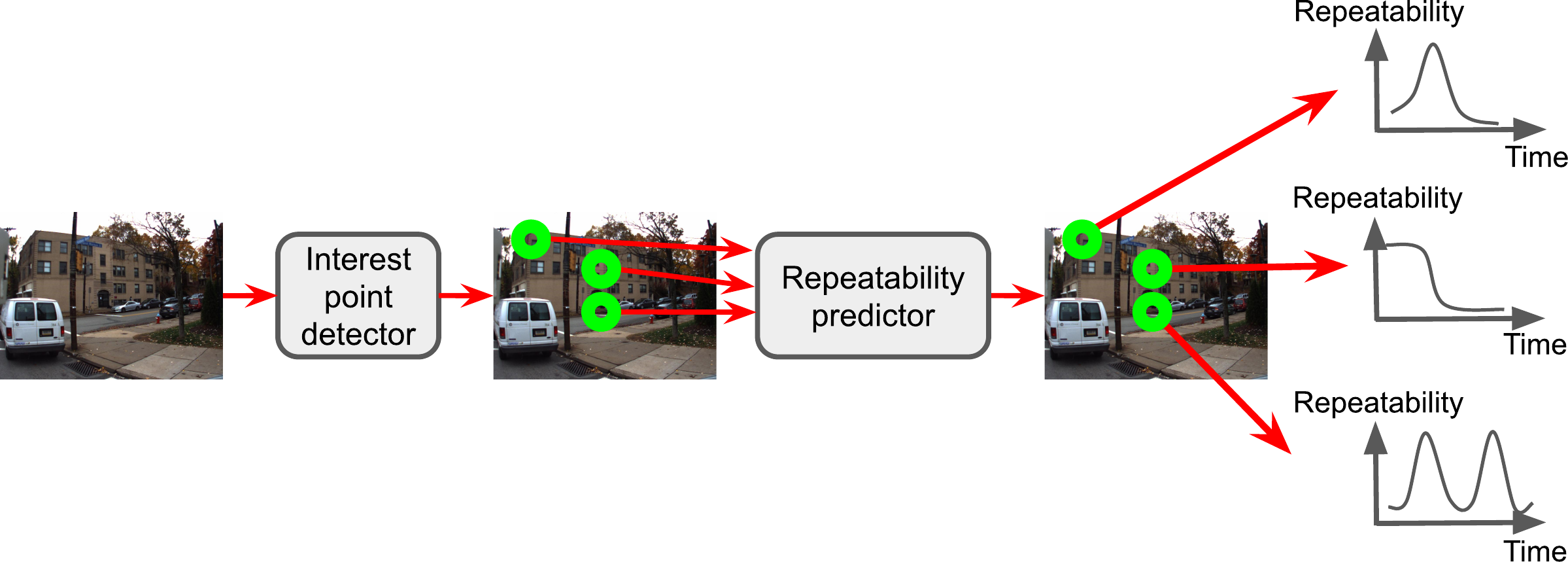}
	\caption{Given an interest point detector, our repeatability predictor predicts the repeatability of interest points (denoted as \tikzcirclegreen{2pt}) as a function of time.}
	\label{fig:concept}
    \vspace{-0.5em}
\end{figure}

Another strategy (e.g.,~\cite{dayoub2008adaptive,dymczyk2016will,maplab}) is to maintain only stable interest points while eliminating others unable to be re-detected. It works efficiently in static indoor environments where interest points are mostly stable for a long period of time. However, in many outdoor environments where dynamic objects exist under varying lighting conditions over days or seasonal changes over years, many interest points likely disappear in some period of time and reappear later. So, a total elimination of them does not keep periodically repeatable interest points.

Such issues motivate us to rethink about the repeatability prediction: \textit{``can we predict if an interest point is repeatable at a particular time period?"}; thereby given a timestamp, we can select appropriate interest points for a certain task at the moment. Our idea is illustrated in Fig.~\ref{fig:concept}, where a given set of interest points detected by a feature detector (e.g., \cite{sift,orb,lift,d2net,r2d2,superpoint}), we predict the repeatability of an interest point as a function of time. In this paper, with the aim of addressing this question, we make two contributions:
\begin{itemize}
\item A repeatability predictor (RP) is formulated as a deep neural network regressor, which receives an interest point, timestamp, and its coordinates as the input, and outputs a vector of repeatability scores approximating the repeatability function (Sec.~\ref{sec:RP}). The ground truth data is built from images periodically captured at multiple viewpoints over time. As alluded above, repeatability should be used in conjunction with matchability to identify if two interest points correspond to a single 3D point, thus we derive the repeatability score at a particular timestamp from the matchability (Sec.~\ref{sec:contruct_ground_truth}).
\item The learned repeatability predictor is applied to the map summarization under visual localization framework~\cite{dymczyk2015gist,dymczyk2015keep,burki2018map}. The main challenge in map summarization lies in the sampling strategy of 3D points---a suboptimal sampling solution will lead to a major degradation of localization accuracy. As our RP is capable of predicting which 3D points are potentially repeatable at the query timestamp, its application can alleviate the accuracy degradation of the map summarization (Sec.~\ref{sec:app_map_sum}).
\end{itemize}
We train our RP on Webcam Clip Art~\cite{webcamclipart} and Extended CMU Seasons~\cite{benchmarkingVL} and show predicting the repeatability for unseen interest points is a feasible task. Additionally we show that by applying RP to map summarization, compared to the baseline, RP significantly prevents the deterioration of map summarization in visual localization accuracy.

\section{RELATED WORK}
To the best of our knowledge, this is the first work to show predicting repeatability of interest points as a function of time is a promising approach, and further apply it to the map summarization for visual localization. 
\subsection{Exploring the properties of interest points}

Since the early work~\cite{schmid98repeatability}, there have been several works in hand-crafted~\cite{sift,orb,surf,brisk} and data-driven approaches~\cite{spencer2020same,tilde,r2d2,cieslewski2019sips,zhai2019learning}, which aim to detect repeatable interest points. As alluded, these approaches assume interest points repeatable w.r.t \textit{all} environmental changes, which usually does not hold in reality due to the physical appearance changes of 3D point~\cite{stylianou2015characterizing}. It is experimentally confirmed by recent benchmarkings~\cite{zhou2016evaluating,VLrevisited}. In particular, as shown by~\cite{stylianou2015characterizing}, the increase in time difference leads to an inevitable failure of feature detection (repeatability), which in turn causes the degradation of matchability. Therefore, \cite{zhou2016evaluating,VLrevisited} show a significant degeneration of feature matching w.r.t severe environmental changes.

Orthogonal methods~\cite{hartmann2014predicting,kim2015predicting,dymczyk2016will} build classifiers to only select stable interest points for certain tasks (e.g., localization). In fact, presuming interest points stable or unstable is equivalent to a binary repeatability function in time, so our work can be viewed as a generalization of those.

\subsection{Map summarization for visual localization}
In large-scale visual localization, the scale of 3D map is too large to fit into a mobile device or even into a modern computer, thus map summarization is necessary not only for dealing with such a large map, but also for reducing the inference time. \cite{minimalscene} formulates the problem to $K$-cover algorithm, which selects a minimal subset of 3D points such that each database image sees at least $K$ number of points regardless of its description. \cite{hybridscenecompression} uses weighted $K$-cover algorithm, which further considers the discriminative power of 3D point descriptors. Instead of decimating points for space coverage, \cite{dymczyk2015keep, maplab} sample 3D points w.r.t observation frequency and low uncertainty. Yet, none of existing work examine timestamp as an input in the map summarization stage. Compared to the baseline~\cite{minimalscene}, we show that timestamp is an another promising constraint for map summarization.

\section{LEARNING REPEATABILITY PREDICTOR}

\begin{figure}
	\centering
	\includegraphics[width=0.48\textwidth]{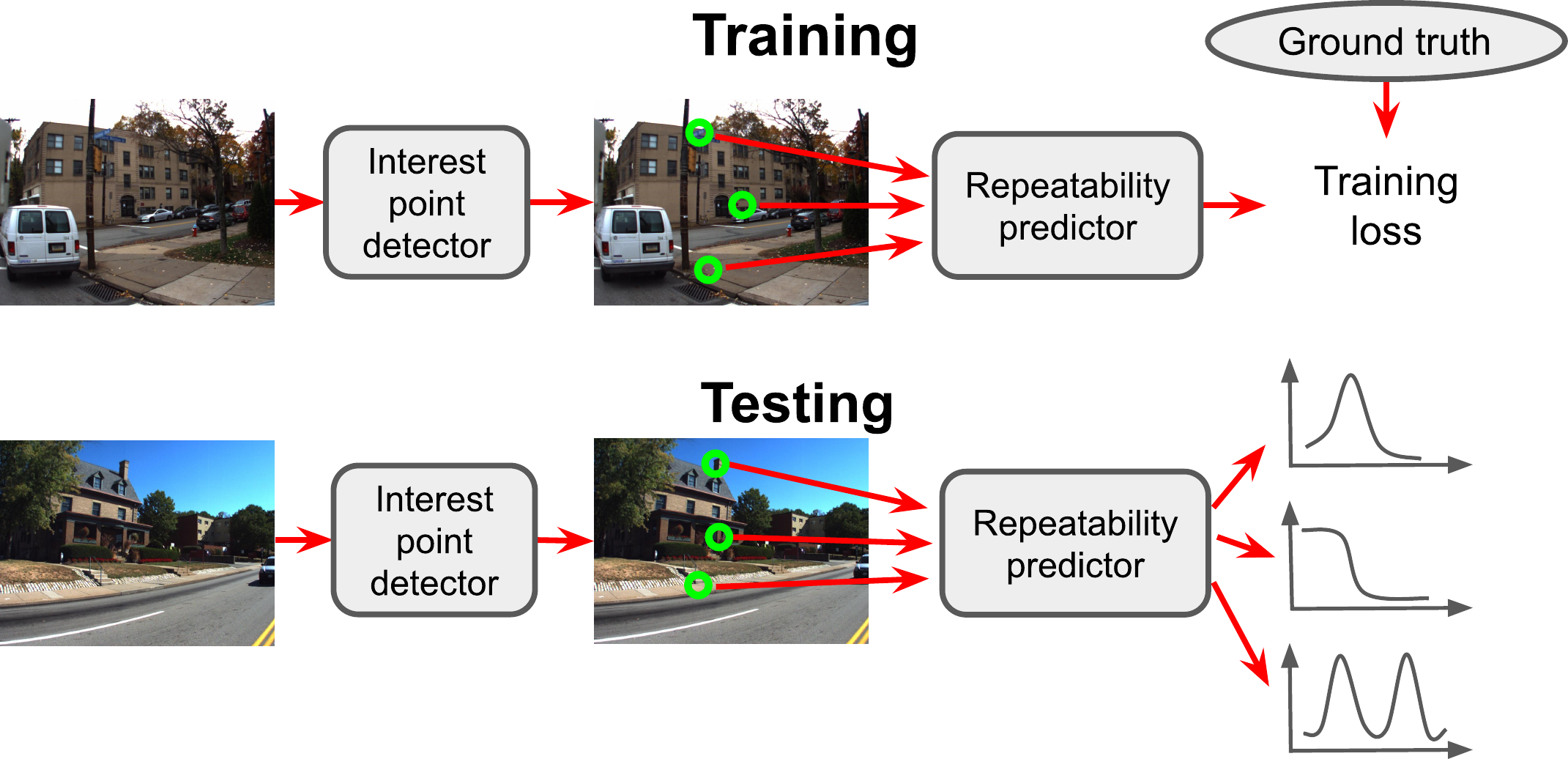}
	\caption{The proposed system for learning the repeatability of the interest points.}
	\label{fig:general_pipeline}
\end{figure}

The proposed system for learning the repeatability of the interest points is shown in Fig.~\ref{fig:general_pipeline}. Given an image, a feature detector extracts its interest points, each interest point is fed to the RP. During the predictor training, the RP is supervised by the ground truth repeatability function. The trained model is then utilized to predict the repeatability function of every interest point for the given test images. In the following sections, we will describe how we parameterize repeatability functions (Sec.~\ref{sec:parameterize_repeat_func}), formulate RP as a deep neural network (Sec.~\ref{sec:RP}), define training loss (Sec.~\ref{sec:loss}), and build the ground truth (Sec.~\ref{sec:contruct_ground_truth}).

\subsection{Parameterizing repeatability functions} \label{sec:parameterize_repeat_func} 

\begin{figure}
	\centering
	\includegraphics[width=0.48\textwidth]{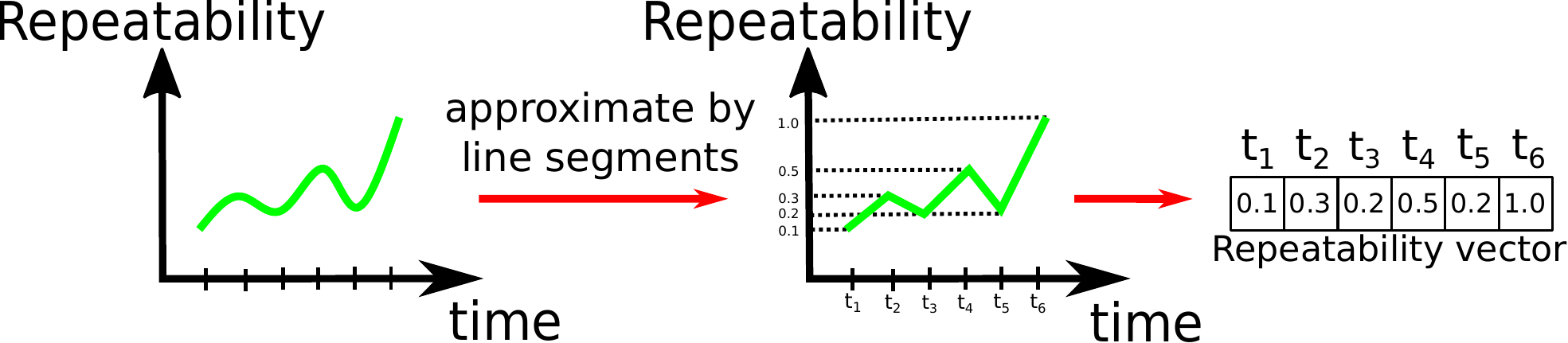}
	\caption{Approximating repeatability function.}
	\label{fig:parameterize_repeatability_function}
\end{figure}

As shown in Fig.~\ref{fig:parameterize_repeatability_function}, we represent a repeatability function over time with a discrete set of line segments. So, time is discretized into timestamps, then line segments are used to approximate the function between two consecutive timestamps. At each timestamp $t_j$, a repeatability score is stored as one element of a vector. Finally, we obtain the repeatability vector which approximates the repeatability function for the given time window.

Let $\bigtriangleup t$ be the interval between two consecutive timestamps and $T$ be the number of timestamps. In different scenarios (see Sec.~\ref{sec:exp_datasets}), we can define $T$ as one of
\begin{itemize}
	\item $T$ = 24 hours per day, $\bigtriangleup t$ = 1 hour, then: $t_1$ = 0:00, $t_2$ = 1:00, $\dots$, $t_{24}$ = 23:00.
	\item $T$ = 365 days per year, $\bigtriangleup t$ = 1 day, then: $t_1$ = January 01, $t_2$ = January 02, $\dots$, $t_{365}$ = December 31. 
	
	\item $T$ = 8760 (= 24$\times$365) hours per year, $\bigtriangleup t$ = 1 hour, then: $t_1$ = 0:00, January 01, $t_2$ = 1:00, January 02, $\dots$, $t_{8760}$ = 23:00, December 31. 
\end{itemize}

\subsection{Repeatability predictor} \label{sec:RP}

\begin{figure*}[ht]
	\centering
	\includegraphics[width=0.80\textwidth]{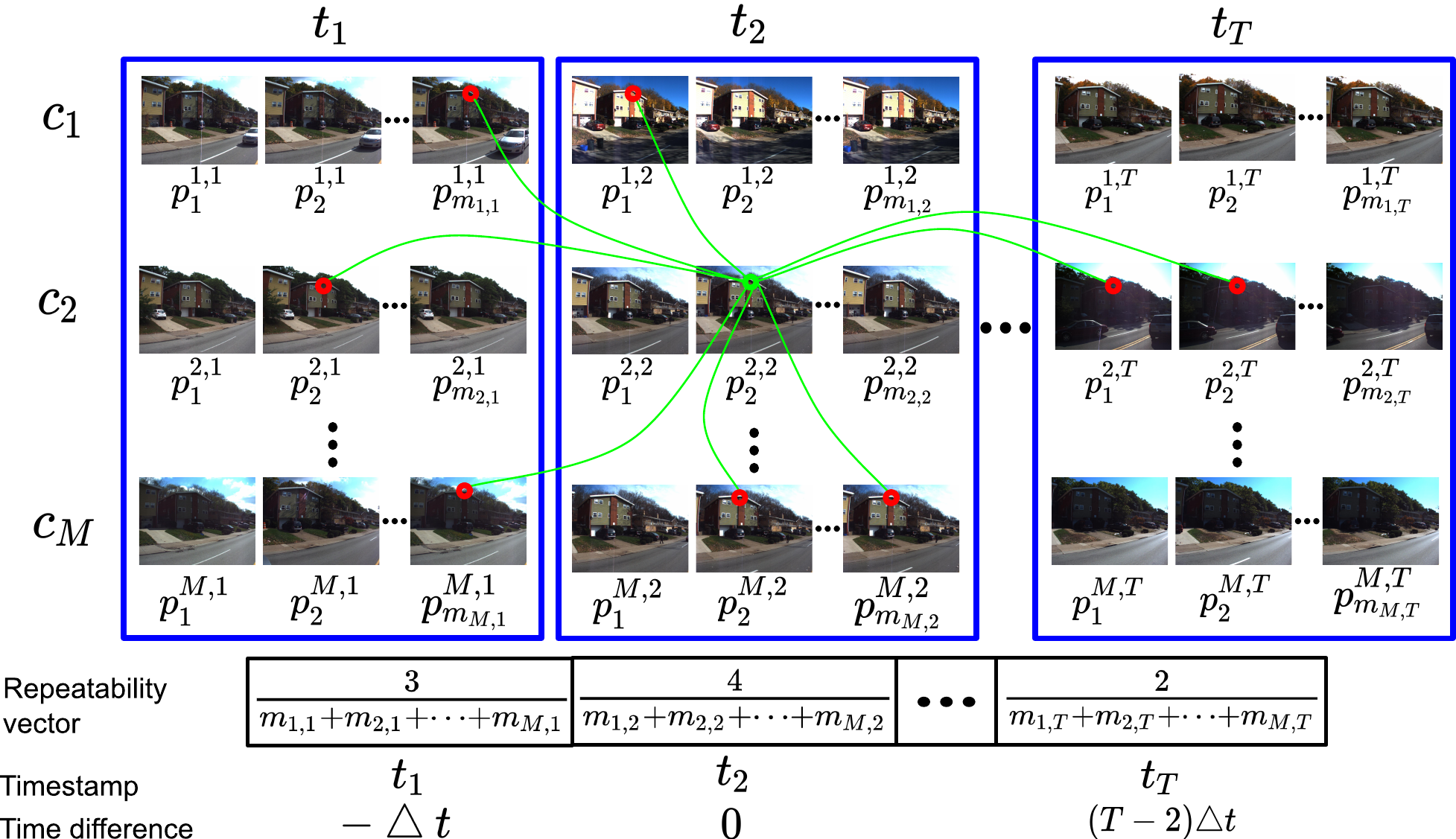}
	\caption{Constructing repeatability vector for the interest point \tikzcirclegreen{2pt}: the repeatability score at timestamp $t_j$ is computed as $\frac{\text{number of inliers \tikzcirclered{2pt}}}{\text{number of images}}$ within the corresponding block \tikzrectangleblue{2pt} of $t_j$.}
	\label{fig:construct_ground_truth}
\end{figure*}

\begin{figure}
	\centering
	\includegraphics[width=0.43\textwidth]{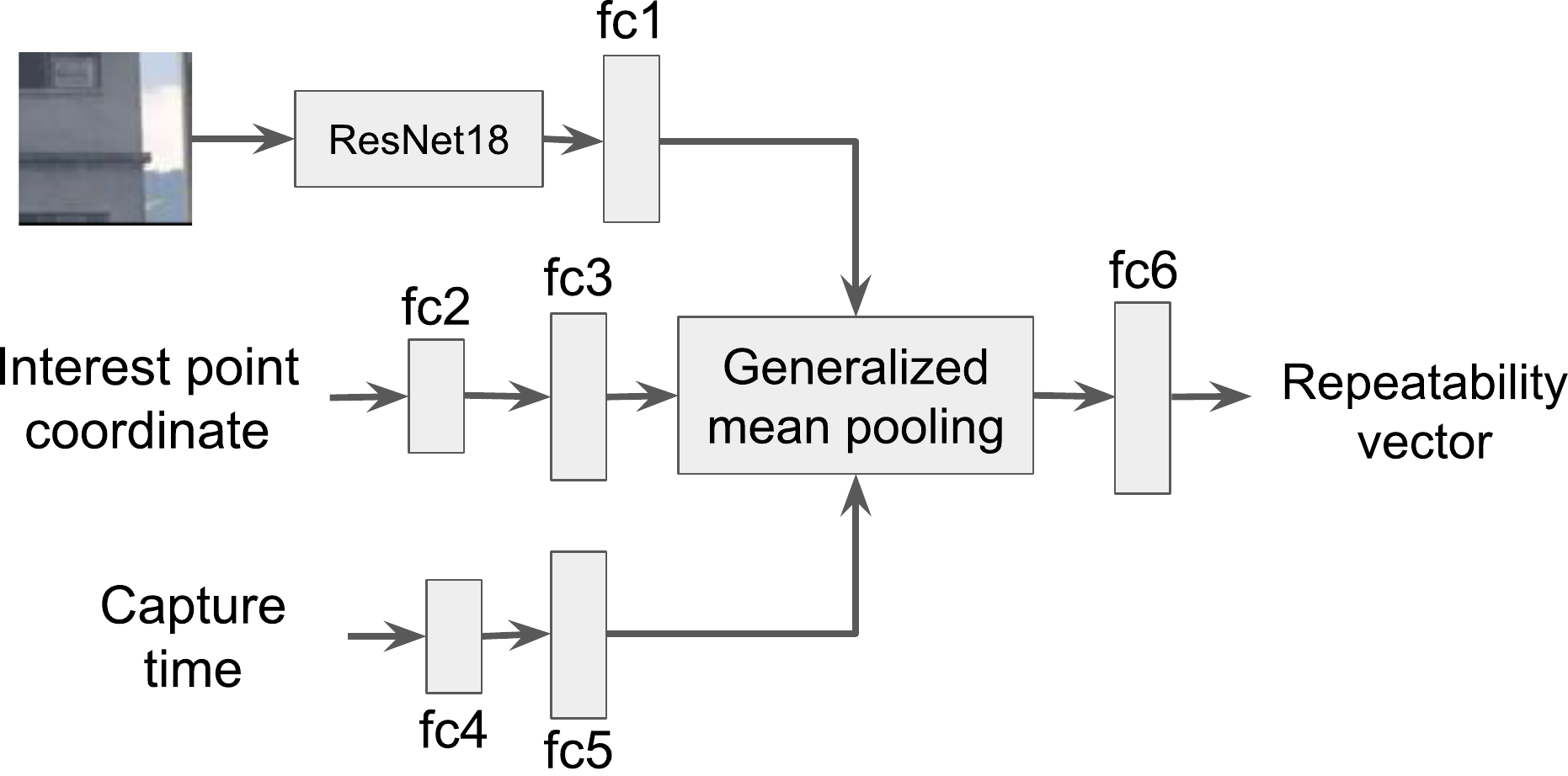}
	\caption{Architecture of repeatability predictor.}
	\label{fig:repeatability_predictor}
	\vspace{-0.5em}
\end{figure}

A deep neural network architecture, as described in Fig.~\ref{fig:repeatability_predictor}, is used to construct the repeatability predictor (RP). For each interest point $x$ in an image, the input of RP consists of the interest point's coordinate ($\in \bbR^2$), local patch ($\bbR^{64\times64\times3}$) centered at the interest point, and the time ($\bbR^{d}$) in which the image is captured, where:
\begin{itemize}
	\item $d=2$ if we represent times of day (hour and minute) or days of year (date and month)
	\item $d=4$ if we represent hours of year (hour, minute, date and month).
\end{itemize}
The information of fully connected layers are: fc1 (64, relu), fc2 (32, relu), fc3 (64, relu), fc4 (32, relu), fc5 (64, relu), fc6 ($D$, sigmoid), where $D$ is the dimensionality of repeatability vector. Outputs of fc1, fc3, and fc5 (denoted as $z_1$, $z_2$, and $z_3$) are pooled by the generalized mean pooling~\cite{generalized_mean_pooling}:

\[
\bz = \begin{bmatrix}\frac{1}{3} \begin{pmatrix} z_1^p + z_2^p + z_3^p \end{pmatrix}\end{bmatrix} ^ {\frac{1}{p}},
\]
where $p$ is a learnable parameter. If $p \rightarrow \infty$, it will become the max pooling; if $p = 1$, it will become the mean pooling.

\subsection{Training loss} \label{sec:loss}
Given $N$ pairs $\begin{Bmatrix} x_i, y_i \end{Bmatrix}_{i=1}^N$ and $T$ timestamps, for a given interest point $x_i$, $y_i = \begin{bmatrix} y^1_i, y^2_i, \dots, y^T_i \end{bmatrix}$ denotes its corresponding ground truth, where $y^j_i \in \mathbb{R}$ is the repeatability score at timestamp $t_j$. Similarly, $\hat{y}_i = \begin{bmatrix} \hat{y}^1_i, \hat{y}^2_i, \dots, \hat{y}^T_i \end{bmatrix}$ denotes the corresponding prediction using RP. The neural network is trained by mean squared error: $\frac{1}{N} \sum_{i=1}^{N} || y_i - \hat{y}_i||_2^2$

\subsection{Constructing ground truth} \label{sec:contruct_ground_truth}

The network with the aforementioned loss function is trained with the ground truth $y$ for the interest point $x$. Fig.~\ref{fig:construct_ground_truth} illustrates the way of generating $y$. For a given set of images viewing the local area (e.g., building, house) captured at $t_1, t_2, \dots, t_T$, the images are grouped into difference cycles $c_1, c_2, \dots, c_M$, where one cycle corresponds to one day (for $T$=24 hours/day) or one year (for $T$=365 days/year or $T$=8760 hours/year). Each image $I_i$ is associated with a camera pose $p_i^{k,j}$ at cycle $c_k$ and timestamp $t_j$. We also denote $m_{k,j}$ as the number of camera poses (images) at cycle $c_k$ and timestamp $t_j$.

For each interest point $x$, we conduct feature matching to each remaining image as follows:
\begin{enumerate}
	\item Find the closest interest point using Euclidean distance between feature descriptors
	\item Verify if the closest interest point satisfies ratio test~\cite{sift}
	\item Conduct geometric verification to check if the matching pair is an outlier
	\item Perform Structure from motion (SfM) and accept the matching pair if it can form a 3D point.
\end{enumerate}
After obtaining set of inlier correspondences, the repeatability score $y^j$ at timestamp $t_j$ is calculated as follows:
\[
y^j  = \frac{\text{\# of inliers}}{m_{1,j} + m_{2,j} + \dots + m_{M,j}}.
\]
Note that if the interest point $x$ belongs to the image captured at timestamp $t_j$, we count itself to the number of inliers. 

Also, our representation in the repeatability function can be seen as \textit{time difference}, e.g., in Fig.~\ref{fig:construct_ground_truth}, let $T$ = 24 hours/day and $\bigtriangleup t$ = 1 hour, interest point $x$ (green point) belongs to the timestamp $t_2$ (= 1:00am), hence the repeatability scores at $y^1$ and $y^T$ are respectively $\bigtriangleup t$ (= 1 hour) before 1:00am and $(T-2) \bigtriangleup t$ (= 22 hours) after 1:00am.

\section{APPLICATION IN MAP SUMMARIZATION} \label{sec:app_map_sum}

\begin{figure}
	\centering
	\includegraphics[width=0.45\textwidth]{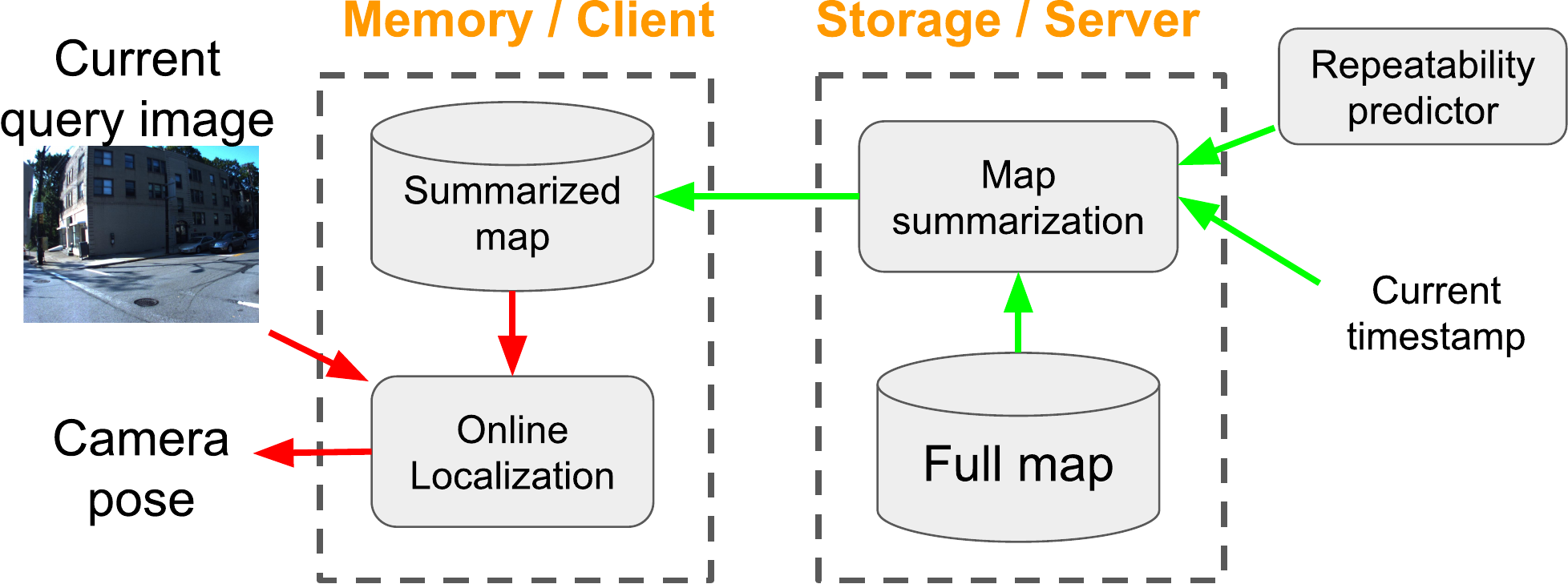}
	\caption{Our proposed VL pipeline, where $\mathcolor{myred}{\pmb{\rightarrow}}$ is the online operation, and $\mathcolor{mygreen}{\pmb{\rightarrow}}$ is the periodic operation, which will be in hourly basis if number of timestamps are $T=24$ hours/day, and $\bigtriangleup t=1$ hour.}
	\label{fig:visual_localization_pipeline}
    \vspace{-0.5em}
\end{figure}

One of the potential applications that benefits from the trained repeatability predictor is map summarization for visual localization (VL). The pipeline is described in Fig.~\ref{fig:visual_localization_pipeline}. Specifically, a full 3D map built by SfM is stored in the server. Given the current timestamp, using RP, the map summarization is performed to obtain a summary map, which is then transmitted to the client and used for online localization. If we set $T = 24$ hours/day and $\bigtriangleup t = 1$ hour, the summary map is updated on the hourly basis.

\subsection{Map summarization} \label{sec:summarizing_3d_map}
\subsubsection{3D point representation} For every 3D point, we predict the repeatability functions for all interest points corresponding to the 3D point. Now, the repeatability of the 3D point is computed by the mean of all the repeatability functions for the interest points. Similarly, the mean of feature descriptors of 2D interest points is also used to represent the descriptor of the 3D point. This representation offers a compact way of storing the 3D map by cutting down the memory consumption on the descriptors and repeatabilities of interest points.

\subsubsection{Sampling 3D points} Firstly, we partition the 3D map into several parts, and then individually prune 3D points in each part. This step prevents us from over-pruning 3D points in a particular part of the map, which would impair the localization accuracy in that part. So, for each part of the map, we compute repeatability score $y_i^j$ of every 3D point $\bp_i$ at the query timestamp $t_j$. Finally, we can remove 3D points with lowest repeatability score according to the pruning ratio. 

\begin{figure}
	\centering
	\mbox{
		\subfloat[][]{
			\includegraphics[width=0.35\textwidth]{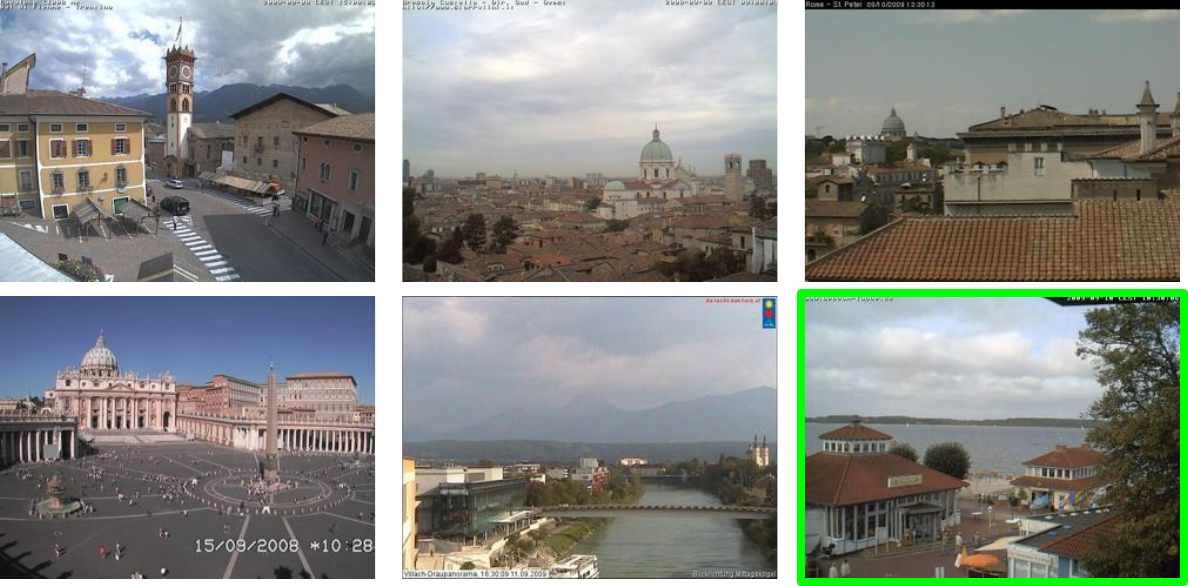}
		\label{fig:data_samples:WebcamClipArt}} 
	}
	
	\mbox{
		\subfloat[][]{
			\includegraphics[width=0.35\textwidth]{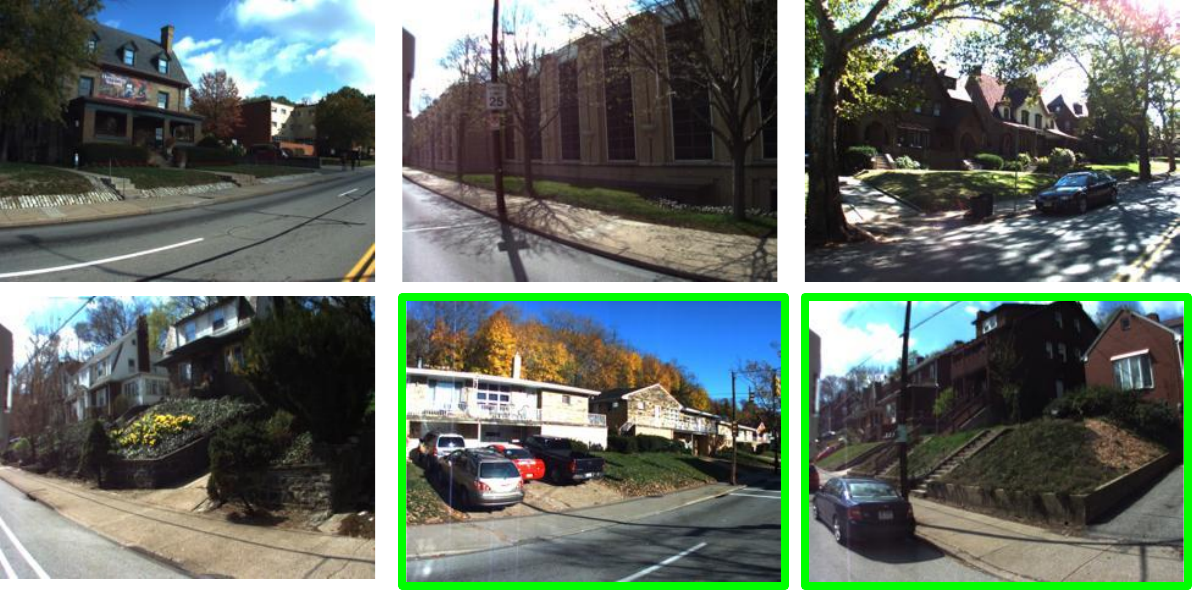}
		\label{fig:data_samples:ExtendedCMUSeasons}}
	}
	\caption{Samples from (a) Webcam Clip Art and (b) Extended CMU Seasons datasets. Testing viewpoints are denoted as \tikzrectanglegreen{4pt}.}
	\label{fig:data_samples}
	\vspace{-0.5em}
\end{figure}

\subsection{Online visual localization}
Given a query image, we firstly retrieve $K$-nearest images in the database. Note that all images are represented by NetVLAD~\cite{netvlad}, and the database images are indexed by $KD$-tree. From the summary map, we select 3D points observed by retrieved images as candidate 3D points. Then, 2D-3D correspondences between interest points of query image and candidate 3D points are established through comparing their feature descriptors and the ratio test \cite{sift} (Note that candidate 3D points are also indexed by $KD$-tree). Finally, the $6$ DoF camera pose of query image is estimated via solving Perspective-n-Point with RANSAC.

\section{EXPERIMENTS}
\subsection{Predicting repeatability function}
This section investigates the performance of our algorithm on several datasets, including Webcam Clip Art dataset~\cite{webcamclipart} and Extended CMU Seasons dataset~\cite{benchmarkingVL}. We use SIFT detector \& descriptor~\cite{sift} for the experiments.

\subsubsection{Datasets} \label{sec:exp_datasets} Two datasets are used:

\begin{figure}
	\centering
	\mbox{
		\subfloat[][]{
			\includegraphics[width=0.40\textwidth]{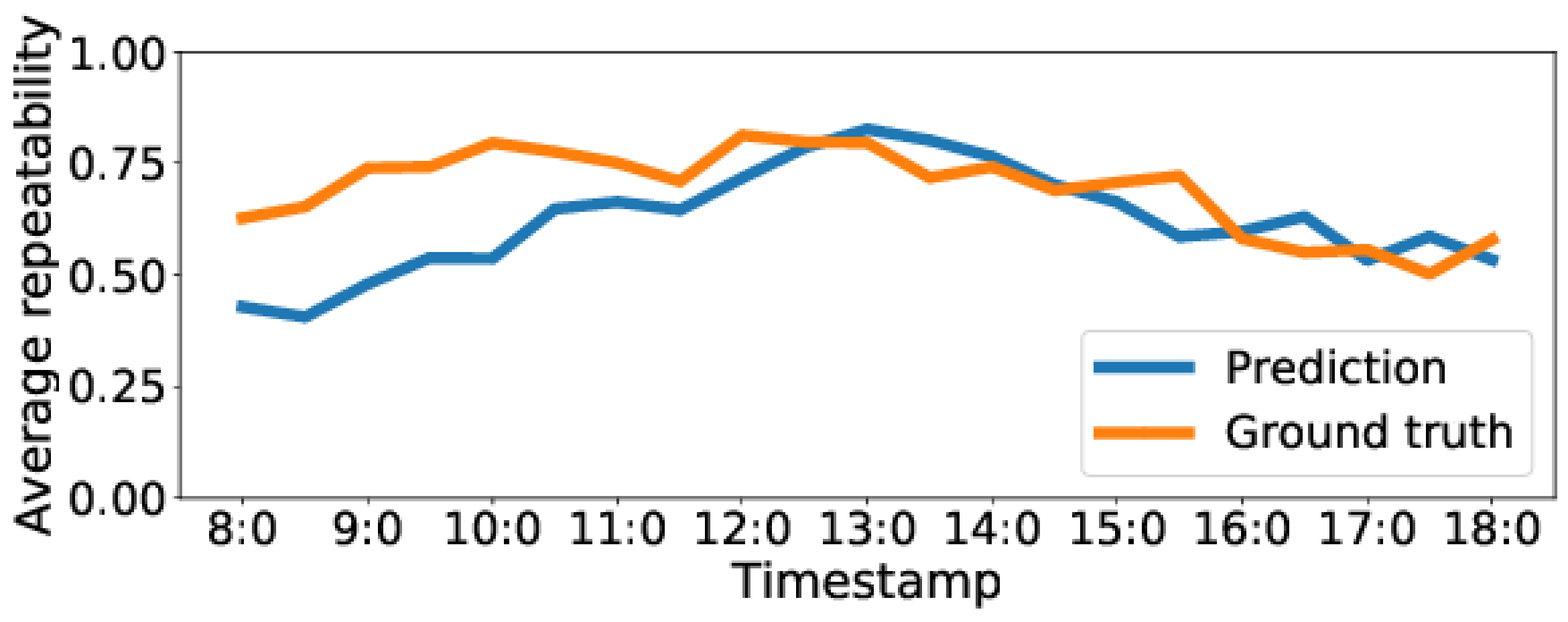}
		\label{fig:average_repeatability_WebcamClipArt}}
	
	}
    \mbox{
    	\subfloat[][]{
		\includegraphics[width=0.40\textwidth]{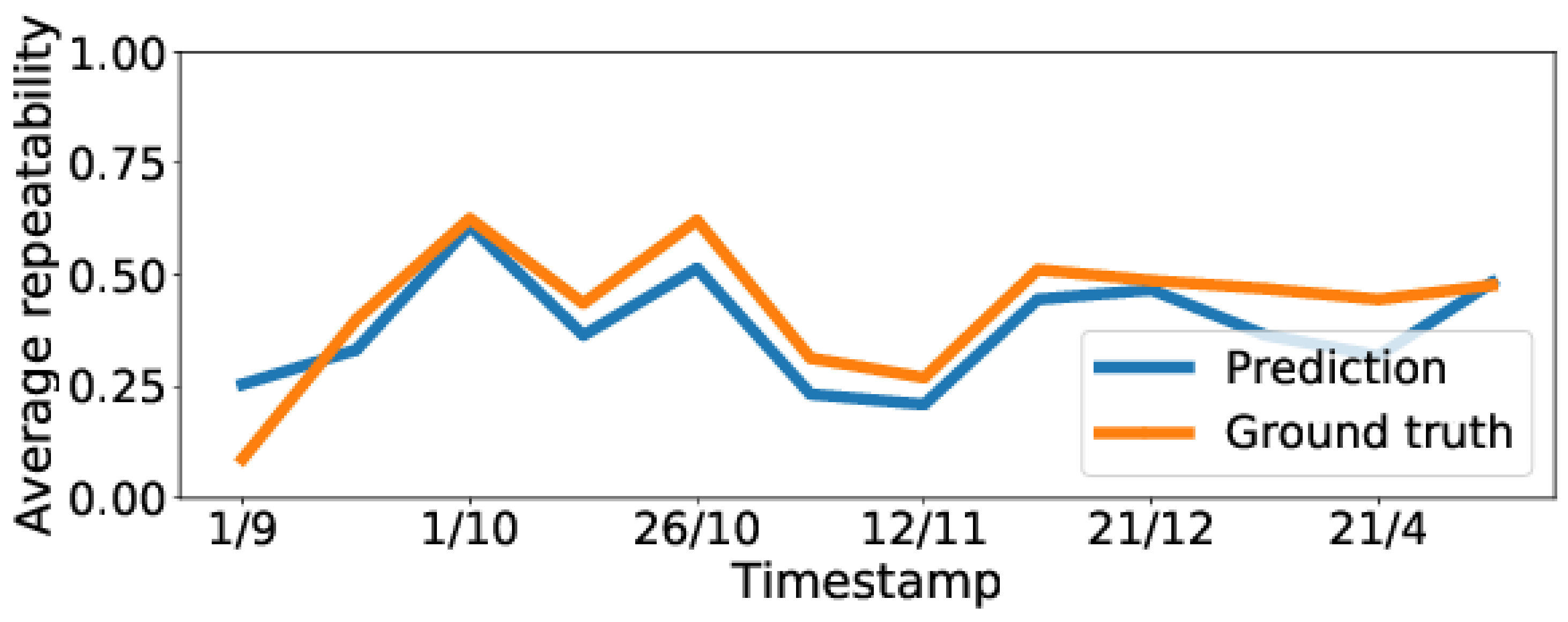}
		\label{fig:average_repeatability_ExtendedCMUSeasons}}
    }

	\caption{Average repeatability over all testing samples on (a) Webcam Clip Art and (b) Extended CMU Seasons.}
	\label{fig:average_repeatability}
  `\vspace{-0.5em}
\end{figure}

\begin{itemize}
	\item Webcam Clip Art~\cite{webcamclipart} has 54 difference viewpoints captured by webcam cameras during several years, and each viewpoint has about 10,000 images. Among the urban viewpoints, we manually select $8$ viewpoints of buildings or houses, which are split to 7 training and 1 testing viewpoints (see samples in Fig.~\ref{fig:data_samples:WebcamClipArt}).
	
	We set the number of timestamps to $T$ = 21 hours/day and $\bigtriangleup t$ = 30 minute between every consecutive timestamps, i.e., $t_1$ = 8:00, $t_2$ = 8:30, $\dots$, $t_{21}$ = 18:00. The number of cycles is $M$ = 4 days. Because the webcam cameras are almost static, we make a minor change in the matching procedure (see Sec.~\ref{sec:contruct_ground_truth}):  in the last step, if the absolute difference between two pixel coordinates is $<$ 5 pixel, the matching pair is accepted as an inlier correspondence. Finally, we obtain 72,160 training and 4,217 testing interest points.
	
	\item Extended CMU Seasons dataset~\cite{benchmarkingVL} is split to 25 separate regions with available ground truth 6 DoF camera poses from SfM. As the dataset is collected in the form of continuous trajectory, we group the images seeing the same viewpoint using available ground truth camera poses.
	In particular, region numbers 6, 7 and 9 are used for training; region 8 is used for testing. After grouping viewpoints, we have 73 training viewpoints and 17 testing viewpoint (see samples in Fig.~\ref{fig:data_samples:ExtendedCMUSeasons}). We set the number of timestamps to $T=12$ days/year and $\bigtriangleup t$ varies from 1 to 13 weeks, i.e., $t_1$ = March 04, $t_2$ = April 21, $t_3$ = July 28, $t_4$ = September 01, $t_5$ = September 15 , $t_6$ = October 01, $t_7$ = October 19, $t_8$ = October 26, $t_9$ = November 03, $t_{10}$ = November 12, $t_{11}$ = November 22, and $t_{12}$ = December 21. The number of cycle is $M$=1 year. In the feature matching in Sec.~\ref{sec:contruct_ground_truth}, due to ground truth camera poses available, we simply perform triangulation instead of the full SfM, resulting in 45,870 training and 15,856 testing interest points.
\end{itemize}

\begin{figure*}
	\centering
	\mbox{
		\subfloat[][]{
			\includegraphics[width=0.25\textwidth]{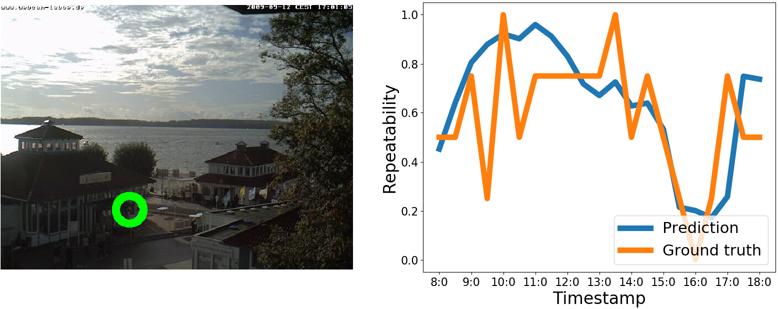}} \hspace{1em}
		\subfloat[][]{
			\includegraphics[width=0.25\textwidth]{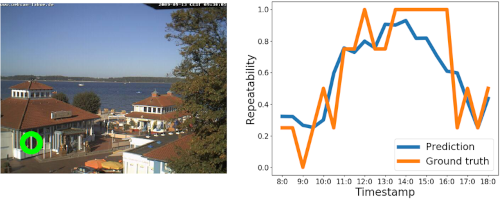}} \hspace{1em}
		\subfloat[][]{
			\includegraphics[width=0.25\textwidth]{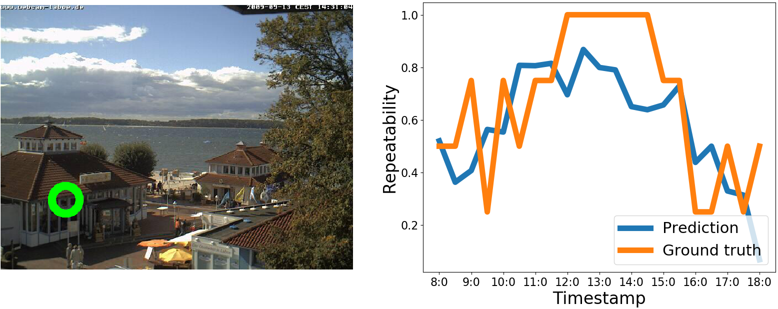}} 
	}
	
	\mbox{
		\subfloat[][]{
			\includegraphics[width=0.25\textwidth]{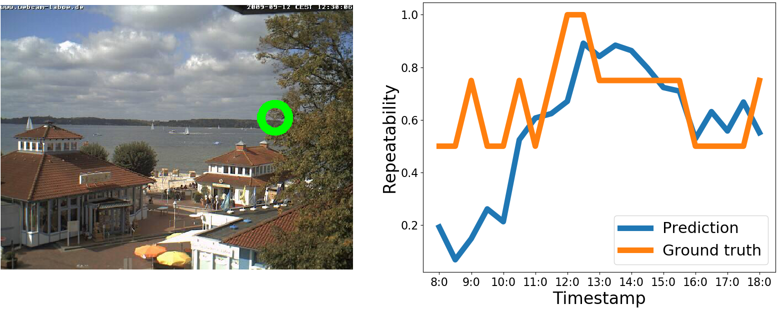}} \hspace{1em}
		\subfloat[][]{
			\includegraphics[width=0.25\textwidth]{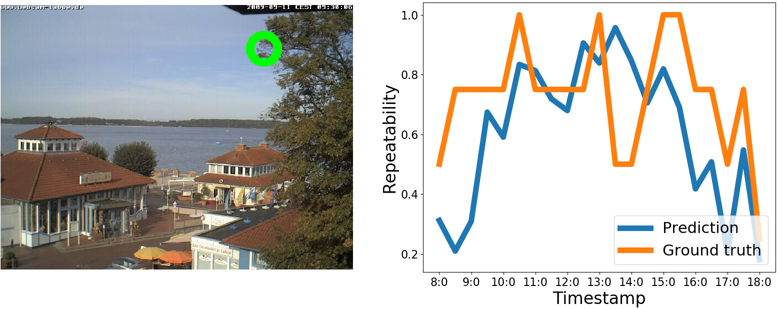}} \hspace{1em}
		\subfloat[][]{
			\includegraphics[width=0.25\textwidth]{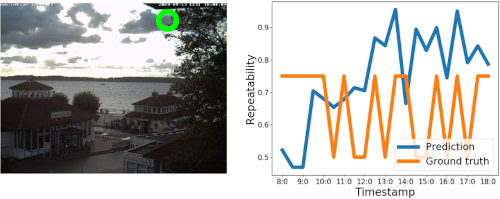}} 
		
	}
	
	\caption{Examples of RP predictions on testing set of Webcam Clip Art, where \tikzcirclegreen{2pt} is the interest point.}
	\label{fig:wla}
	\vspace{-0.35em}
\end{figure*}

\subsubsection{Results} Fig.~\ref{fig:average_repeatability_WebcamClipArt} shows the average repeatability function over all testing samples on Webcam Clip Art. Generally, in both ground truth and prediction curves, the repeatability score increases from the morning to noon, and gradually decreases as it gets close to the night time. This trend can also be seen in individual testing samples (see Fig.~\ref{fig:wla}a-e). However, for testing sample with unclear ground truth trend (see Fig.~\ref{fig:wla}f), RP struggles to learn its repeatability.

\begin{figure*}
	\centering
	\mbox{
		\subfloat[][]{
			\includegraphics[width=0.25\textwidth]{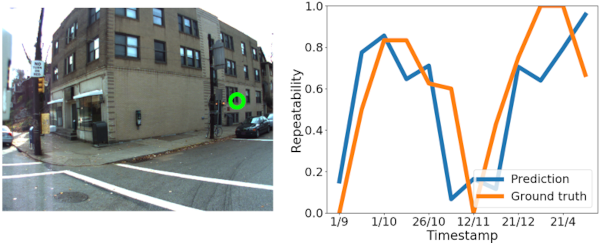}} \hspace{1em}
		\subfloat[][]{
			\includegraphics[width=0.25\textwidth]{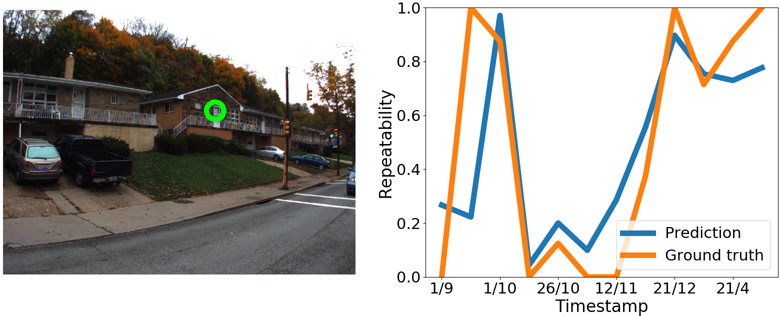}} \hspace{1em}
		\subfloat[][]{
			\includegraphics[width=0.25\textwidth]{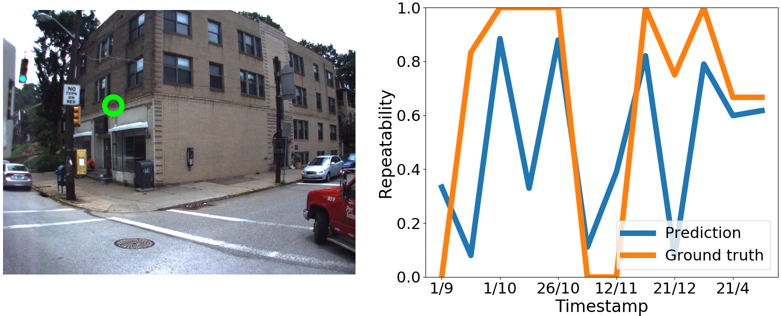} \label{fig:ecs:fail_building}}
	}

	\mbox{
		\subfloat[][]{
			\includegraphics[width=0.25\textwidth]{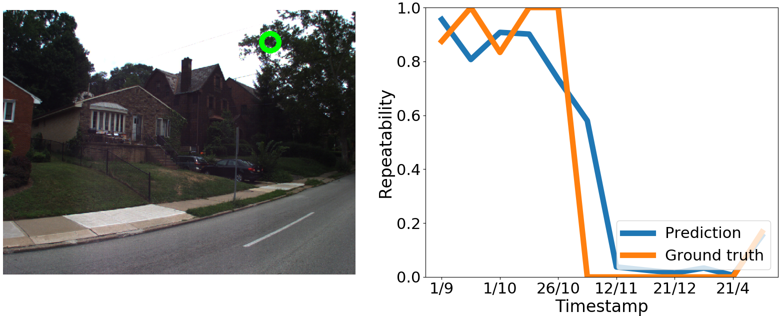} \label{fig:ecs:success_tree}} \hspace{1em}
		\subfloat[][]{
			\includegraphics[width=0.25\textwidth]{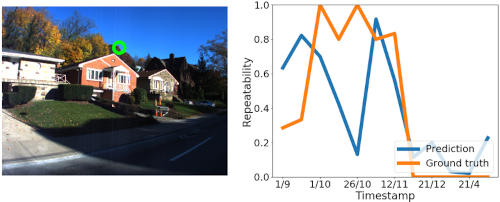} \label{fig:ecs:fail_tree}}  \hspace{1em}
		\subfloat[][]{
			\includegraphics[width=0.25\textwidth]{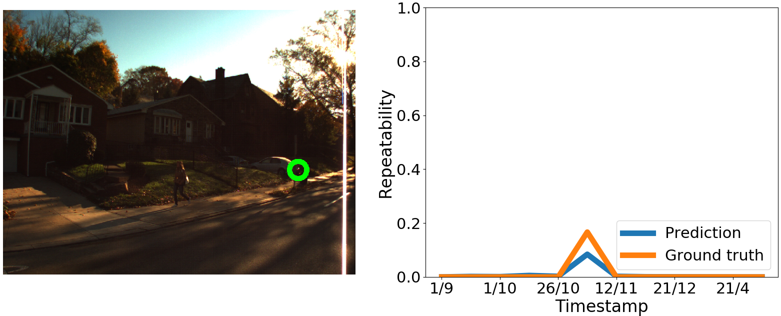} \label{fig:ecs:success_car}} 
	}
	
	\caption{Examples of RP predictions on testing set of Extended CMU Seasons, where \tikzcirclegreen{2pt} is the interest point.}
	\label{fig:ecs}
	 \vspace{-0.5em}
\end{figure*}
\begin{figure*}[h]
	\centering
	\mbox{
		\subfloat{
			\includegraphics[width=0.29\textwidth]{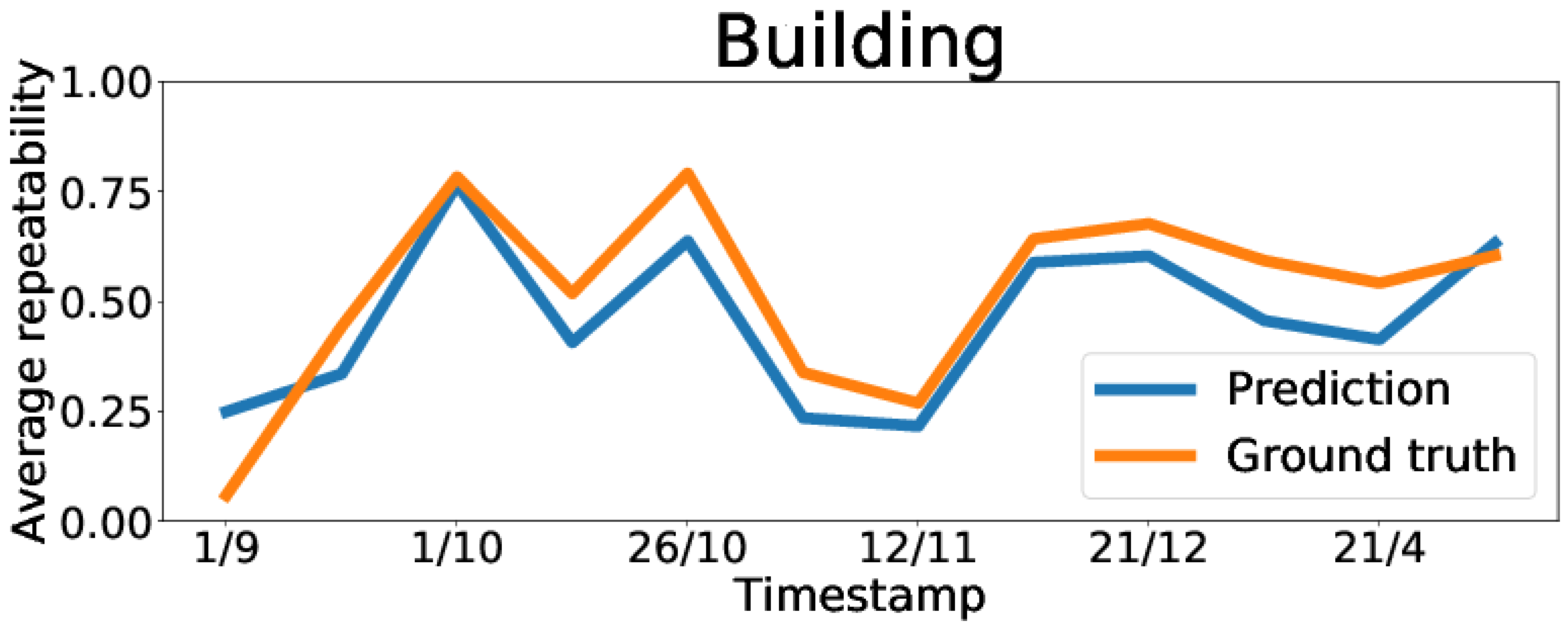}}  \hspace{1em}
		\subfloat{
			\includegraphics[width=0.29\textwidth]{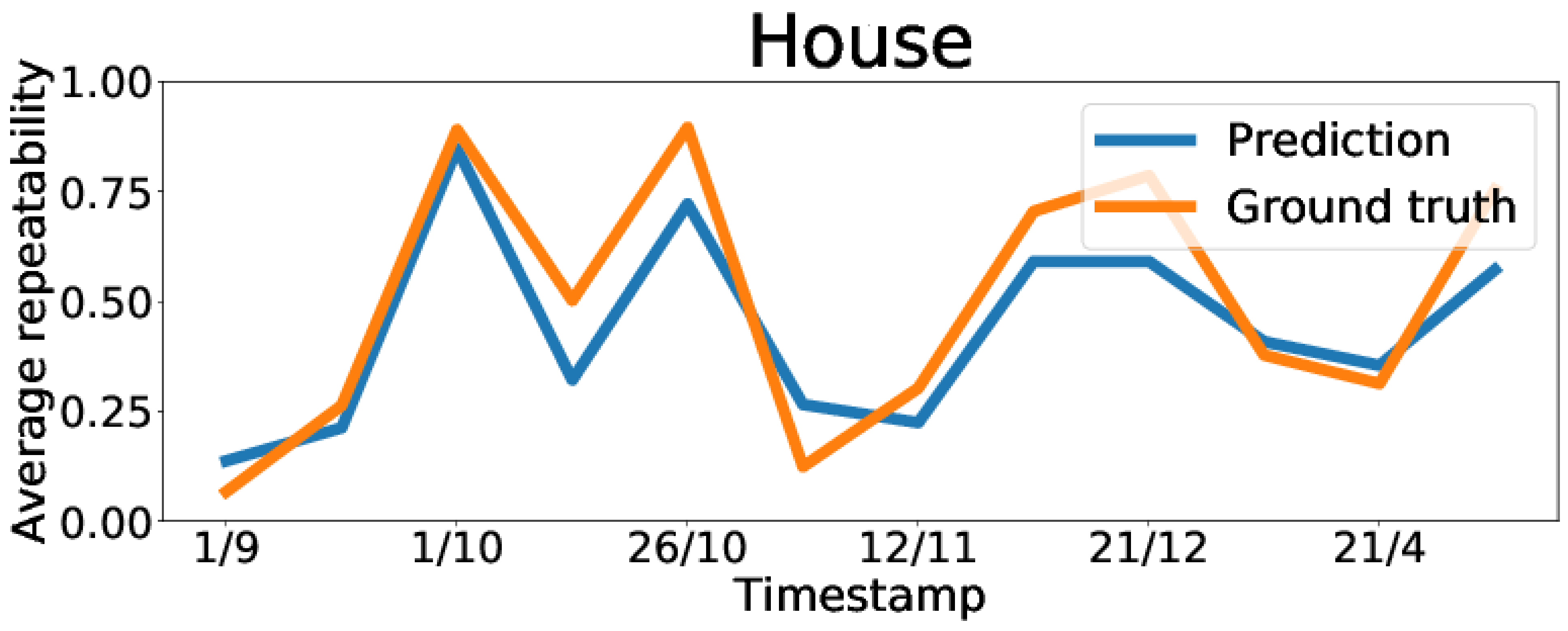}}  \hspace{1em}
		\subfloat{
			\includegraphics[width=0.29\textwidth]{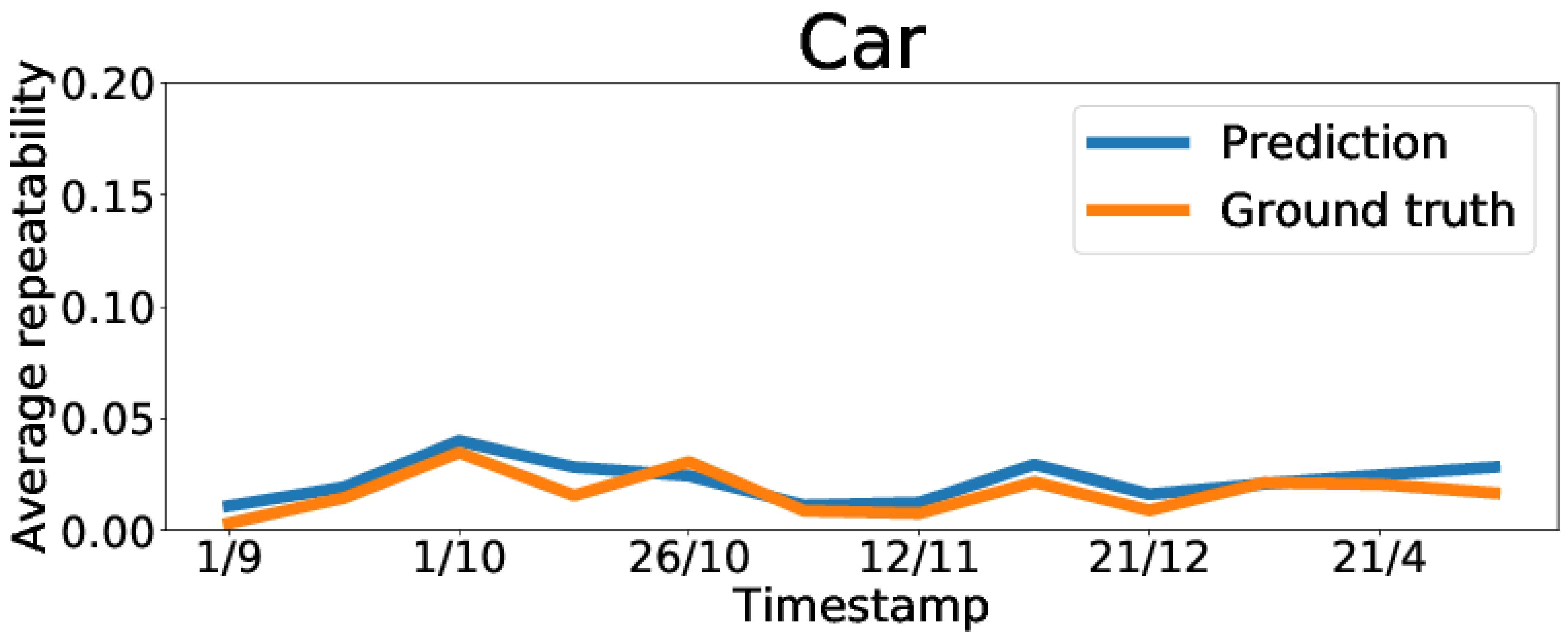}}	
		}

	\mbox{
		\subfloat{
			\includegraphics[width=0.29\textwidth]{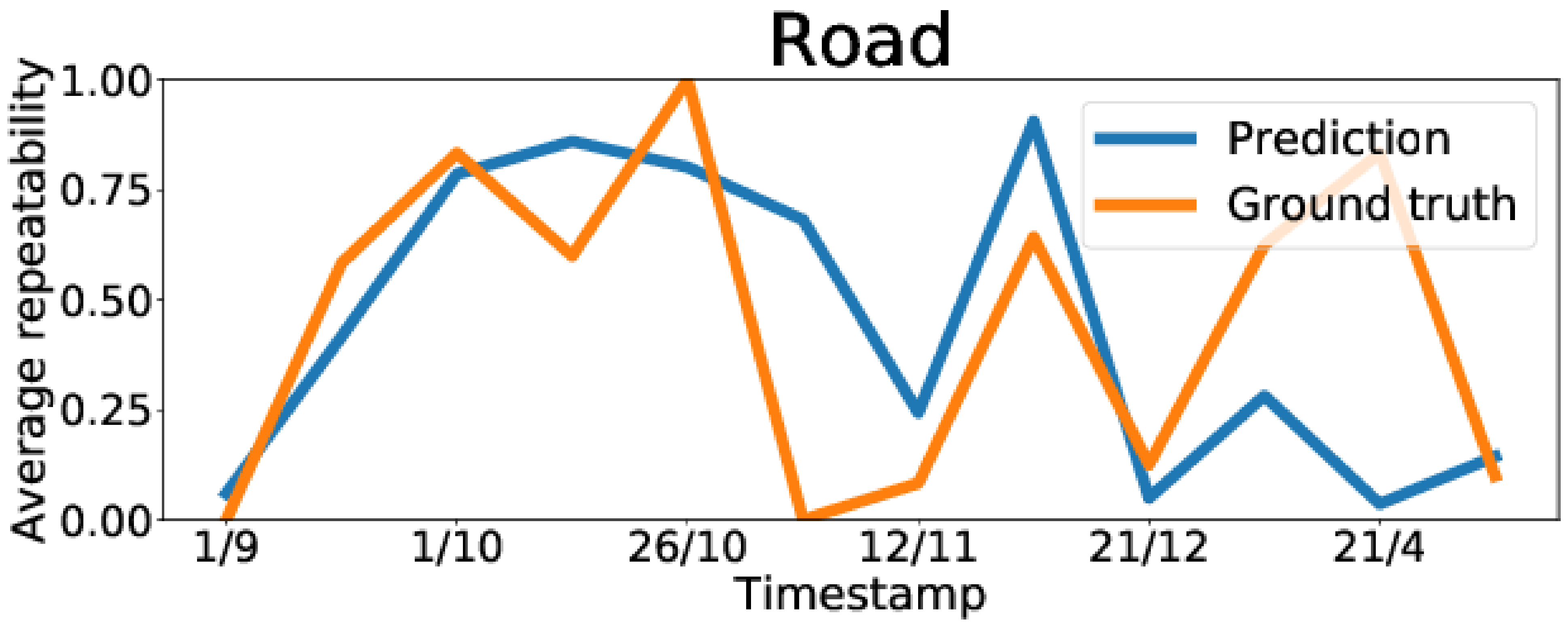}}  \hspace{1em}
		\subfloat{
			\includegraphics[width=0.29\textwidth]{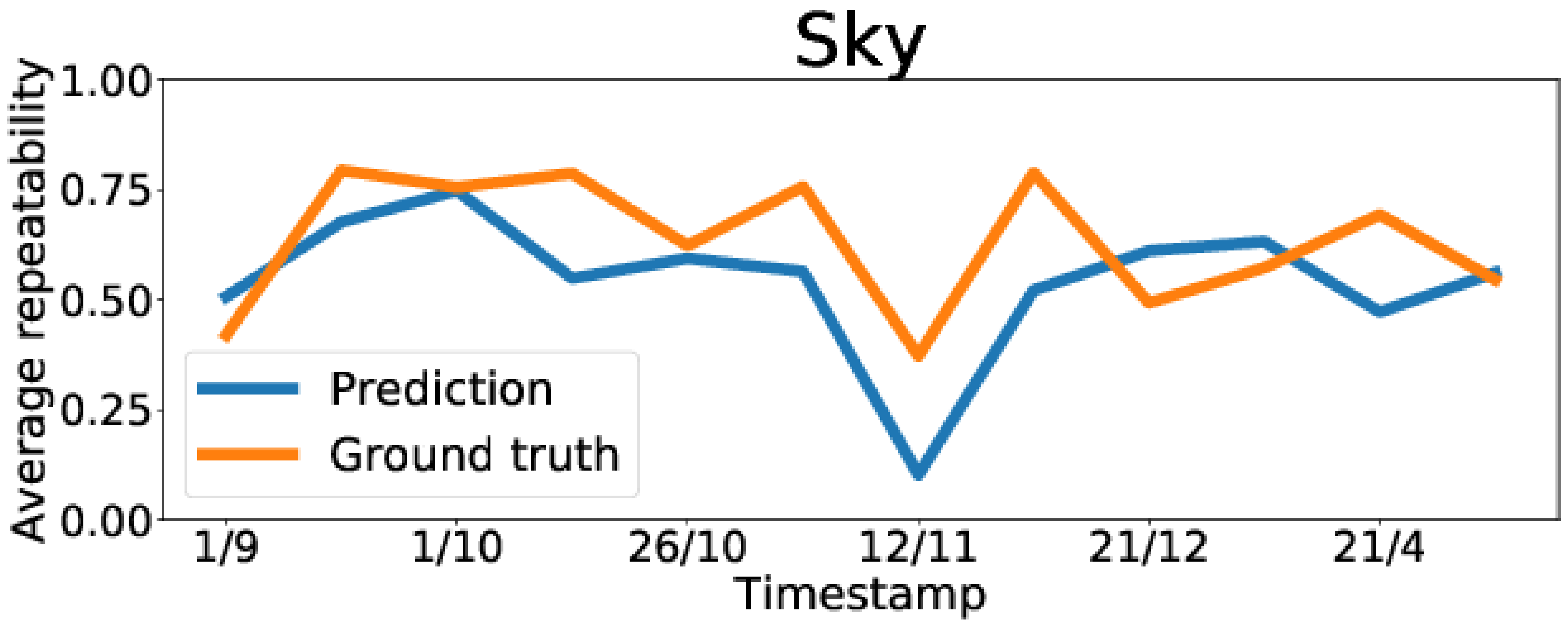}}  \hspace{1em}
		\subfloat{
			\includegraphics[width=0.29\textwidth]{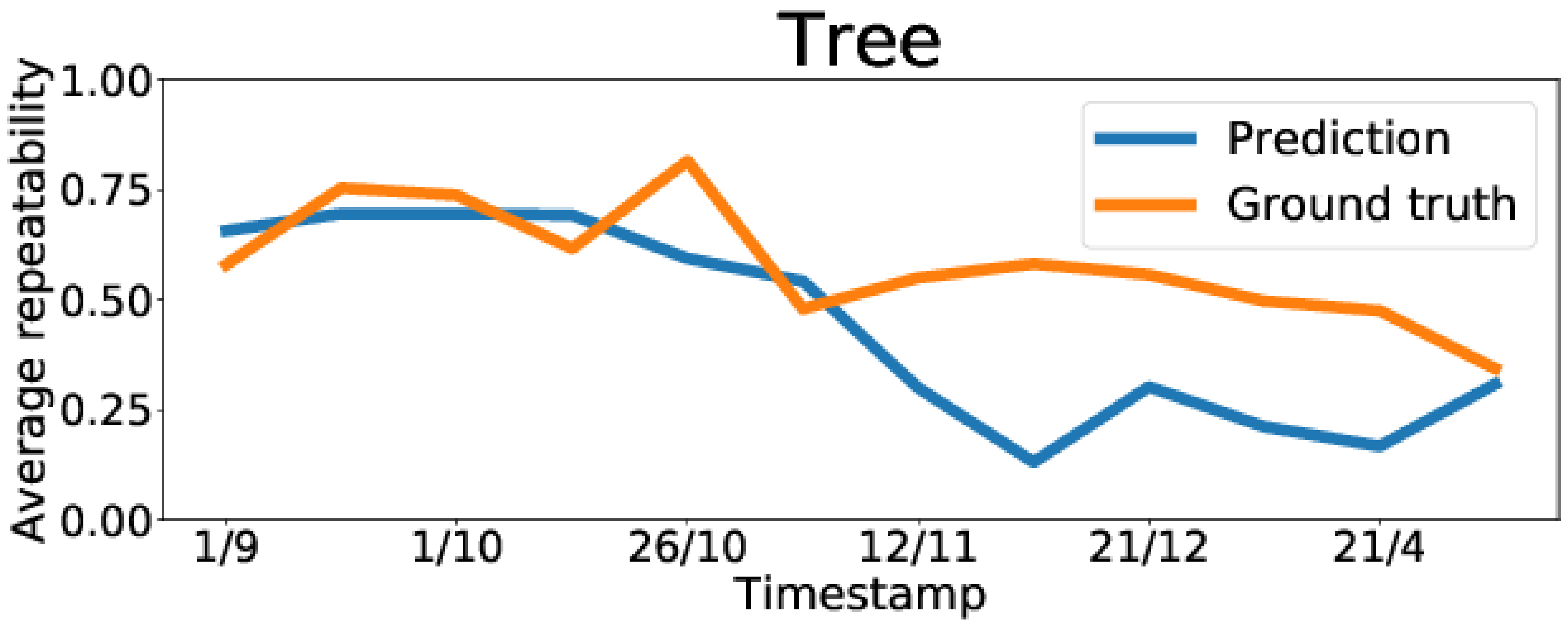}}		
	}

	\caption{Average repeatability in each category over all testing samples on Extended CMU Seasons.}
	\label{fig:average_repeatability_each_category}
    \vspace{-0.5em}
\end{figure*}

Fig.~\ref{fig:average_repeatability_ExtendedCMUSeasons} shows the average repeatability function over all testing samples on Extended CMU Seasons. It is clear that the ground truth and the prediction curves share a similar trend, i.e., the repeatability score is high in spring, summer and autumn while it is low during winter. This trend mainly comes from interest points of discriminative objects, e.g., buildings, houses (see Fig.~\ref{fig:average_repeatability_each_category}), which RP can perform prediction reasonably. The concrete good examples are shown in Fig.~\ref{fig:ecs}a-b, but RP also fails in few examples (e.g., Fig.~\ref{fig:ecs:fail_building}). 

For tree interest points, the basic trend is their repeatability drastically drops after winter (due to no leaves after winter), which RP often fails in predicting (see Fig.~\ref{fig:average_repeatability_each_category}). Fig.~\ref{fig:ecs:success_tree} and  Fig.~\ref{fig:ecs:fail_tree} show the good and bad particular examples. 

Interest points of dynamic objects (e.g., cars) are mostly repeatable at the capture time, i.e., the input timestamp (see Fig.~\ref{fig:ecs:success_car}), yielding a good prediction in general (see Fig.~\ref{fig:average_repeatability_each_category}). 

For background interest points (e.g., road, sky), their repeatability shows an unclear trend, thus RP struggles to perform a reasonable performance (see Fig.~\ref{fig:average_repeatability_each_category}).

Based on the comprehensive experiment, we empirically observe several challenges in this setup: 1) as shown in \cite{stylianou2015characterizing}, sun direction is an important factor affecting the appearance of physical 3D points, which should be considered as an input, 2) Fig.~\ref{fig:average_repeatability_each_category} shows the performance of RP varies according to types of interest points. It suggests that the predictor may benefit from semantics, and 3) another factor which greatly influences to the prediction result is the weather at the query time, e.g., at 12pm with a rainy and cloudy condition, the appearance might be darker than that at 5pm with a sunny and clear condition.

\subsection{Map summarization for VL}
\begin{table}
	\centering
	\caption{Sequences in Oxford RobotCar used to train RP.}

	\begin{tabular}{|c|c|}
	
		\hline
		Sequence & Timestamp \\
		\hline \hline
		2015-02-13-09-16-26 & $t_1 = \text{9:00}$ \\
		2015-07-10-10-01-59 & $t_2 = \text{10:00}$\\
		2015-03-17-11-08-44 & $t_3 = \text{11:00}$ \\
		2014-11-28-12-07-13 & $t_4 = \text{12:00}$ \\
		2014-11-18-13-20-12 & $t_5 = \text{13:00}$ \\
		2015-07-29-13-09-26 & $t_6 = \text{14:00}$ \\
		2015-05-19-14-06-38 & $t_7 = \text{15:00}$ \\
		2015-08-13-16-02-58 & $t_8 = \text{16:00}$ \\
		2015-07-14-16-17-39 & $t_9 = \text{17:00}$ \\
		\hline
		
	\end{tabular}
	\label{tab:oxford_robotcar_seqs}
   \vspace{-0.5em}
\end{table}

\subsubsection{Datasets} Oxford RobotCar dataset~\cite{OxfordRobotCar} is used to train the proposed network. Specifically, we set number of timestamps to $T$ = 9 hours/day (from 9:00 to 17:00), $\bigtriangleup t$ = 1 hour, and number of cycle $M$ = 1. More detailed information of the Oxford RobotCar sequences is described in Table \ref{tab:oxford_robotcar_seqs}. To generate separate viewpoints, we manually select 23 viewpoints by taking the latitude and longitude, and the yaw angle of the vehicle seeing them, which, for convenience, are now denoted as \textit{viewpoint latitude-longitude}, and \textit{viewpoint yaw angle}. Afterward, using the exact ground truth pose of the Oxford RobotCar~\cite{oxford_robotcar_kinematic_groundtruth}, we find images close to each viewpoint latitude-longitude $\leq$ 5m. To ensure images observing the same viewpoint, we only keep images whose yaw angle close to viewpoint yaw angle $\leq$ 45$^\circ$. Finally, SfM is conducted in each viewpoint; and only 18 viewpoints which have all images registered are retained for training RP. 

\begin{figure}
	\centering
	\includegraphics[width=0.40\textwidth]{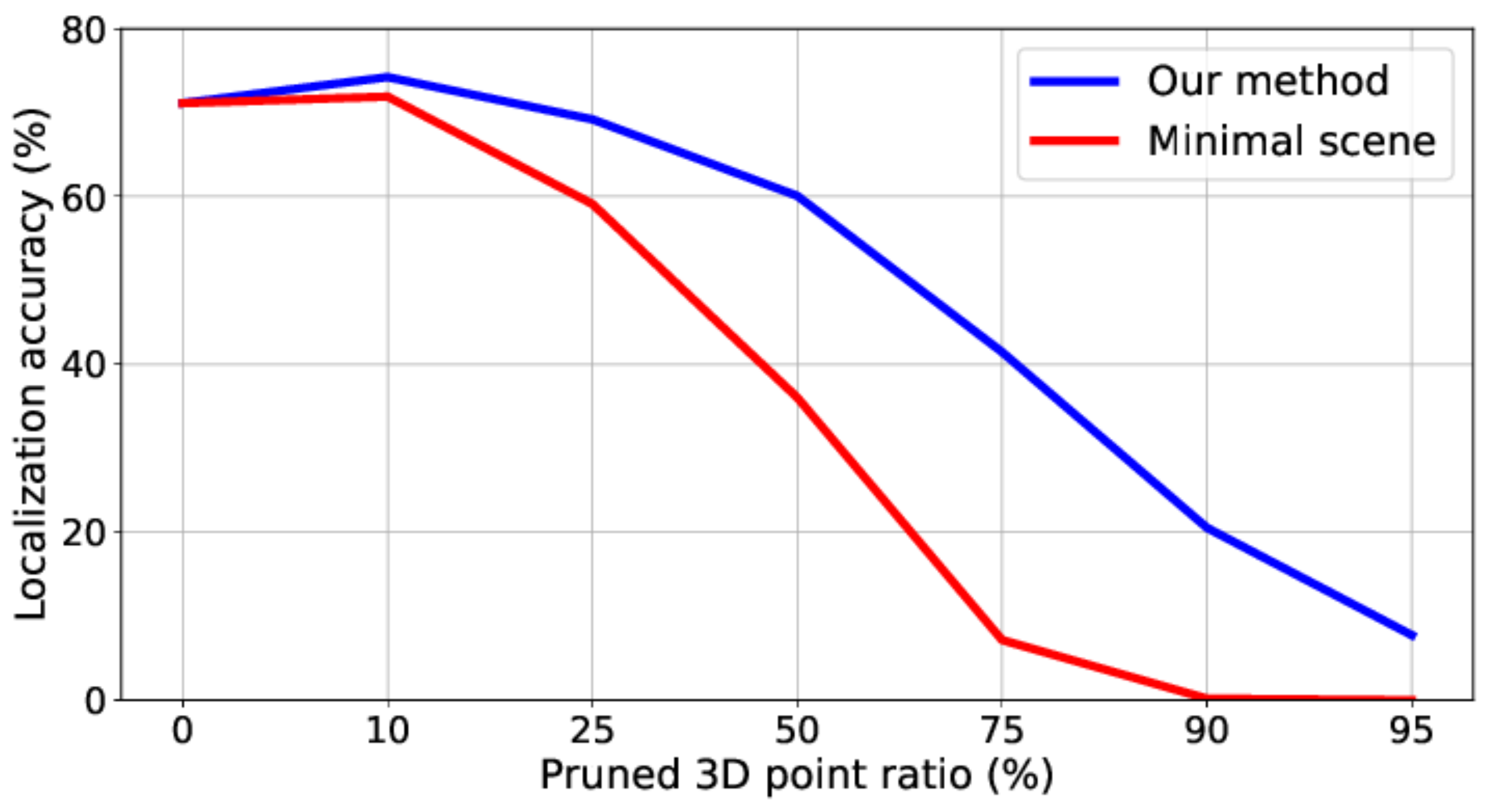}
	\caption{Comparison of localization accuracy over the percentage of pruned 3D points}
	\label{fig:map_sum_result}
    \vspace{-1.2em}
\end{figure}

For VL dataset, we select region 8 of Extended CMU Seasons~\cite{benchmarkingVL}. Its 3D point cloud is used as the 3D map, and query images are captured at 04 March 2011.

\subsubsection{Results} In Extended CMU Seasons, there is a capture timestamp associated with every query image. Utilizing that timestamp and RP trained on Oxford RobotCar, we summarize the 3D map as described in Sec.~\ref{sec:summarizing_3d_map}. In Fig.~\ref{fig:map_sum_result}, where the localization accuracy is the percentage of query images localized $< 5\text{m},10^\circ$, our method shows a superior accuracy to \textit{Minimal scene}~\cite{minimalscene} in every pruned 3D point ratio. The reason is \textit{Minimal scene}~\cite{minimalscene} does not consider the matching potential of 3D points in VL when selecting them, while our method regards the query time to sample highly repeatable 3D points, leading to a better performance. 

\section{CONCLUSION AND FUTURE WORKS}

Interest points have an important role in many robotic vision applications, hence detecting interest points is a vital problem. In contrast to the existing methods that aim to detect permanently repeatable interest points, this paper proposes to train a repeatability predictor (RP) which can predict the repeatability of an interest point as a function of time. Through comprehensive experiments, an insightful analysis is provided, i.e, each type of interest points has a specific trend, enabling RP to predict its repeatability; and the existing challenges in this problem are also discussed. We believe if those are addressed properly, a ``universial" RP could be achieved. Furthermore, an application of RP on map summarization for visual localization (VL) is provided, i.e., RP suggests potentially repeatable 3D points at the query time, then VL framework can sample appropriate 3D points, preventing the deterioration of its performance. The experiment shows its significant potential in the VL problem. In future, because the repeatability decay over a long period of time is inevitable, parameterizing this trend in the repeatability function by basis functions is necessary. A ``universial" RP also requires a large amount of data captured in more cycles and viewpoints. In addition, more applications of RP in robotic vision will be explored.


\bibliographystyle{IEEEtran}
\bibliography{IEEEabrv,egbib}

\end{document}